\documentclass[runningheads]{llncs}
\usepackage[hidelinks]{hyperref}
\usepackage[T1]{fontenc}
\usepackage{graphicx}
\usepackage{amsmath,amssymb,amsfonts}
\usepackage{xcolor}
\usepackage{color}
\usepackage{subcaption}
\usepackage{booktabs}
\usepackage{array}
\usepackage{orcidlink}
\definecolor{lightgray}{gray}{0.9}

\usepackage{xcolor}%
\usepackage{textcomp}%
\usepackage{manyfoot}%
\usepackage{booktabs}%
\usepackage{algorithm}%
\usepackage{algorithmicx}%
\usepackage{algpseudocode}%
\usepackage{listings}%
\usepackage{float}
\usepackage{orcidlink}
\usepackage{balance}
\usepackage{enumitem}
\setlist{nosep}

\begin{document}
\title{Visual Retrieval-Augmented Generation for Silhouette-Guided Animal Art}
\titlerunning{Visual-RAG for Silhouette-Guided Animal Art}
%
\author{
Quoc-Duy Tran\inst{1,2}\orcidlink{0009-0003-1773-1520} \and
Anh-Tuan Vo\inst{1,2}\orcidlink{0009-0002-3547-9907} \and
Trung-Nghia Le\thanks{Corresponding author.}\inst{1,2}\orcidlink{0000-0002-7363-2610}}

\authorrunning{Quoc-Duy Tran et al.}
%
\institute{University of Science, VNU-HCM, Ho Chi Minh, Vietnam \and
Vietnam National University, Ho Chi Minh, Vietnam\\
\email{\{22120082,22120406\}@student.hcmus.edu.vn}\\
\email{ ltnghia@fit.hcmus.edu.vn}}
\maketitle              

\begin{abstract}
Generative AI has advanced the ability to render photorealistic or artistic images, yet it remains limited in a key aspect of human creativity: interpreting ambiguous shapes. This phenomenon, rooted in pareidolia, allows humans to perceive meaningful forms in random patterns such as clouds, stones, or leaves. To computationally replicate this imaginative process, we introduce Visual Retrieval-Augmented Generation (Visual-RAG), a framework that generates animal art directly from natural silhouettes. Our method retrieves structurally similar animal shapes from a curated corpus of 28,586 high-quality silhouettes and uses them as reference exemplars to guide diffusion-based generation with ControlNet and IP-Adapter. Ablation studies confirm that shape Context with RANSAC provides the most accurate alignment, while removing shape standardization reduces the inlier ratio to just 13.4\%, underscoring the importance of structural fidelity in Visual-RAG. A user study with 12 participants evaluated the outputs in terms of aesthetics, silhouette fidelity, and overall impression. Results reveal that while Visual-RAG provides plausible interpretations, challenges remain in achieving high perceptual impact. This work lays the foundation for computational pareidolia, showing how machines can contribute to the early stages of imaginative discovery.

\end{abstract}
%

\vspace{-5mm}

\section{Introduction}\label{sec:intro}


Creating artistic imagery from ambiguous natural forms is a direct exercise in divergent perception, a cognitive ability linked to creativity \cite{Bellemare2022}. This capacity is rooted in pareidolia, a phenomenon considered the basis of iconographic art, where viewers resolve induced ambiguity \cite{Bednarik2016}. 
However, "seeing things for what they could be" is cognitively challenging. As a rapid, subjective, and fleeting shortcut rooted in survival instincts \cite{Bednarik2016}, pareidolia is difficult to sustain for artistic purposes; studies show more creative individuals experience it more often \cite{Bellemare2022}. Bridging this perceptual gap to finished artwork demands both imagination and technical skill.

Modern generative AI, despite its advances, provides limited support for this creative process. Text-to-image models such as Stable Diffusion \cite{Rombach_2022_CVPR} require explicit prompts, burdening the user with conceptualization. Shape-conditioned systems like ControlNet \cite{zhang2023addingconditionalcontroltexttoimage} excel at rendering but cannot autonomously interpret ambiguity, lacking 'appreciation' for their own artifacts \cite{Heath2016}. Meanwhile, existing computational pareidolia approaches \cite{song2021everythingstalkinpareidoliaface,Wan:2022:Cloud2Sketch} remain confined to narrow domains. Current frameworks are thus tools, not creative partners, failing 'perceptually grounded' creativity \cite{Heath2016}.

To address this gap, we propose a computational framework that automates the imaginative interpretation of natural object silhouettes by integrating retrieval with generative. Our visual retrieval-augmented generation, Visual-RAG in short, employs an IP-Adapter to transfer the visual appearance of a retrieved animal image into a new generation while constraining the output to the given silhouette. To support this research, we curated a dataset of 28,586 masked animal instances derived from OpenImagesV7 \cite{OpenImages}, providing a large-scale resource for retrieval-based experimentation. 

We further conducted experiments to evaluate the proposed method. The ablation study demonstrates that reliable geometric alignment is far more important than raw speed for silhouette-guided generation. Shape Context with RANSAC consistently provided the most accurate matches, while shape standardization proved essential for preserving structural fidelity. Together, these components form the foundation that enables Visual-RAG to produce coherent and perceptually meaningful outputs. Meanwhile, the user study focuses on several dimensions of creative quality, including aesthetic appeal, shape fidelity, and overall impression. This study advances perceptually grounded creative AI by introducing a novel framework and revealing key trade-offs between generative and retrieval-based strategies.

Our contributions are as follows:
\begin{itemize}
    \item We introduce Visual-RAG, a computational framework for generating animal art directly from ambiguous silhouettes.

    \item We construct and release a curated dataset of 28,586 masked animal images from OpenImagesV7.

\end{itemize}

\vspace{-5mm}

\section{Related Work}\label{sec:related}

\vspace{-2mm}


\textbf{Shape-based retrieval} has progressed from classical geometric descriptors to modern learning-based methods. Early techniques such as Shape Context (SC) \cite{Belongie:2002:ShaepContext} and Inner-Distance Shape Context (IDSC) \cite{Ling:2007:IDSC} captured boundary and articulation information, while faster descriptors like Hierarchical String Cuts (HSC) \cite{Wang:2014:HSC} traded accuracy for efficiency, often struggling with highly articulated shapes. More recent advances employ convolutional networks \cite{radenović2018deepshapematching} and graph neural networks \cite{monti2016geometricdeeplearninggraphs}, offering learned shape representations but requiring large annotated datasets and often showing limited generalization to naturalistic silhouettes in creative contexts. To address scalability, post-processing strategies such as Online-to-Offline (O2O) \cite{Zheng2019:O2O} shift computational load to offline preprocessing, improving efficiency for large-scale retrieval. Our approach adapts this principle while emphasizing retrieval quality over speed by incorporating robust geometric verification with RANSAC-based alignment. This balance ensures reliable matching performance in the open-ended and diverse scenarios central to creative applications.


\textbf{Conditional image generation} has advanced from early GAN-based methods such as Pix2Pix \cite{isola2018imagetoimagetranslationconditionaladversarial} to diffusion models with fine-grained structural conditioning exemplified by ControlNet \cite{zhang2023addingconditionalcontroltexttoimage}. While these models excel at rendering, they still depend on explicit prompts and lack autonomous interpretative ability. Related paradigms such as content-aware inpainting, demonstrated by EdgeConnect \cite{nazeri2019edgeconnectgenerativeimageinpainting} and DeepFillv2 \cite{yu2019freeformimageinpaintinggated}, show strong contextual reasoning yet remain limited to completing missing regions rather than generating novel interpretations. More recently, IP-Adapter \cite{Ye:2023:IPAdapter} has extended diffusion models with appearance transfer, enabling reference images to guide generation while preserving structural constraints. Building on this capability, our retrieval-based approach transfers animal appearance characteristics into arbitrary silhouettes, bridging perceptual ambiguity with creative synthesis.


Originally introduced in NLP \cite{Lewis:2020:RAG}, \textbf{retrieval-augmented generation (RAG)} integrates parametric knowledge with external retrieval and has been applied to tasks such as few-shot learning \cite{gao:2025:clipadapterbettervisionlanguagemodels} and image captioning \cite{ramos:2023:retrievalaugmentedimagecaptioning}. We extend this idea to creative generation, where retrieved examples act as inspirations that expand interpretative possibilities. This direction echoes "learning from imaginary data" \cite{wang2018lowshotlearningimaginarydata}, but shifts hallucination from refining classification boundaries to enabling creative exploration. Multimodal retrieval has been advanced through Vision-Language Models (VLM) like CLIP \cite{Radford:2021:CLIP}. Building on these foundations, our framework demonstrates how shape-based retrieval can inform semantic interpretation through large language models, supporting perceptually grounded creative synthesis.

\vspace{-5mm}

\section{Proposed Method}\label{sec:methods}

\vspace{-2mm}

\begin{figure}[!t]
 \centering
 \includegraphics[width=\textwidth]{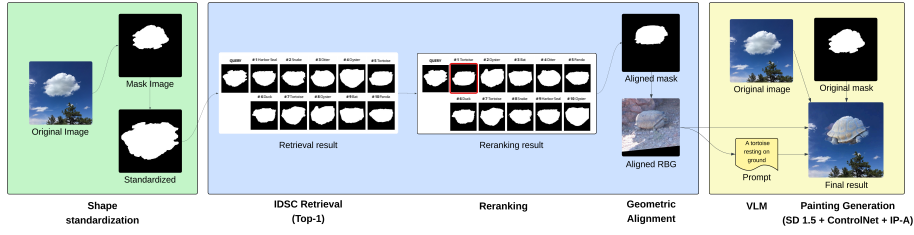}
 \vspace{-4mm}
 \caption{Overview of the visual retrieval-augmented generation pipeline.}
 \label{fig:visual_rag}
 \vspace{-2mm}
\end{figure}

Given a high-quality corpus of animal silhouettes, we aim to leverage this data by finding a structurally compatible instance from the corpus to serve as a detailed visual reference. Figure \ref{fig:visual_rag} illustrates an overview of this approach.

\textbf{Silhouette Segmentation. } We adopt a two-step segmentation pipeline to isolate the target object without manual annotation. Grounding DINO~\cite{Zhang:2023:GroundingDINO} first detects candidate bounding boxes $\mathcal{B} = {b_1, b_2, \ldots, b_n}$ from open-vocabulary prompts (e.g., “stone,” “cloud,” “fire”). The highest-confidence box $b^i$ is then passed to SAM~\cite{kirillov2023segany}, which outputs the final segmentation mask $M$.

\textbf{Geometric Feature Extraction. } Geometric fidelity is critical in this retrieval stage. Although the O2O framework \cite{Zheng2019:O2O} provides an effective balance between speed and accuracy, its reliance on fast descriptors during the online stage is suboptimal for our task. Meanwhile, HSC \cite{Wang:2014:HSC} captures only coarse global structure, which often yields geometrically incongruent matches in our highly articulated animal dataset. Because our generative model depends on structurally coherent references, such mismatches cannot be corrected downstream. To mitigate this issue, we adopt the classic SC descriptor \cite{Belongie:2002:ShaepContext}, which offers articulation-aware matching at the cost of increased computational overhead. This choice prioritizes structural accuracy, ensuring reliable retrieval for subsequent generation.

\begin{figure}[!t] 
    \centering
    \includegraphics[width=0.8\textwidth, height = 5cm]{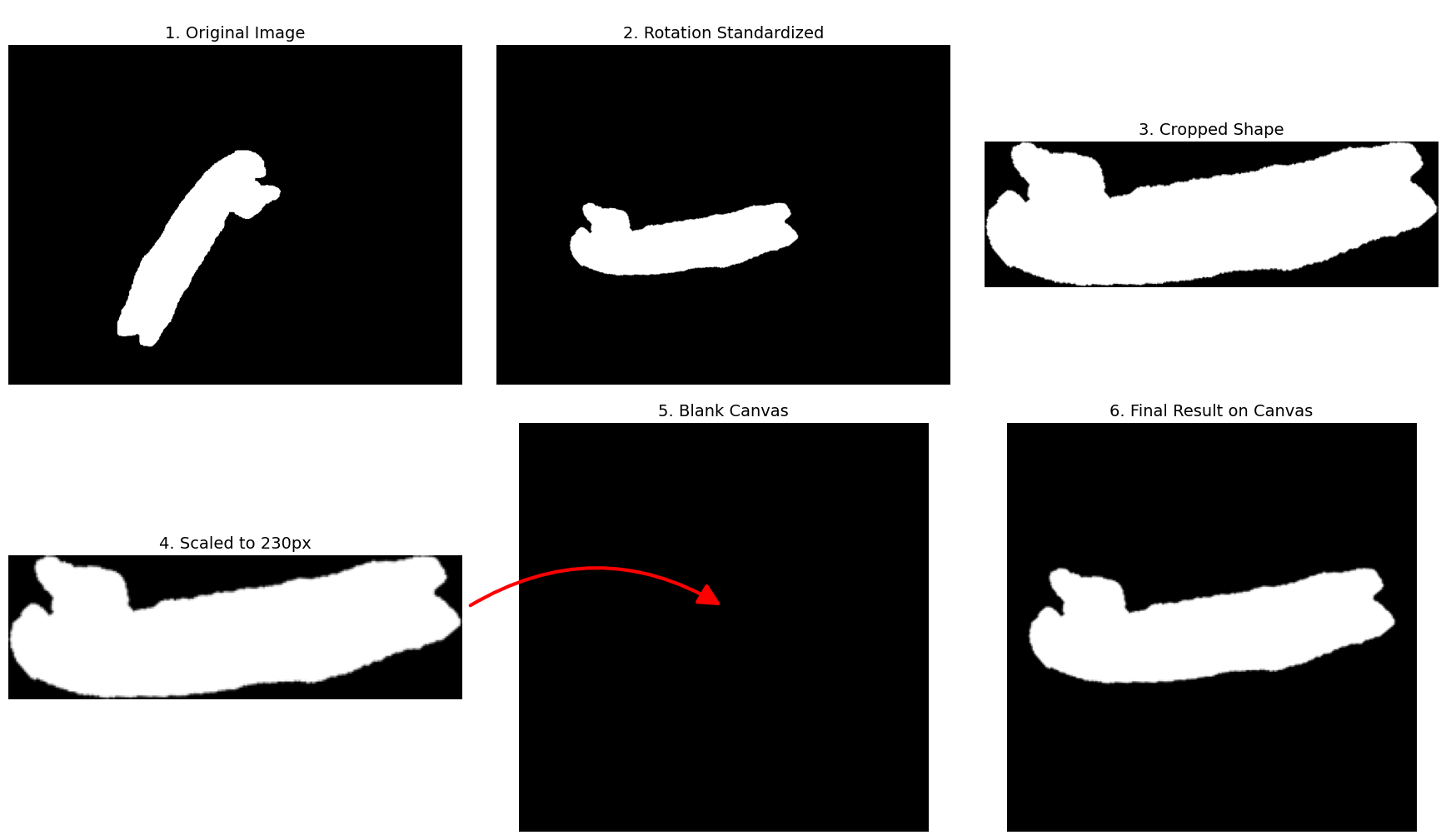}
    \vspace{-2mm}
    \caption{Shape standardization process.
    \textbf{(1)} Original binary mask.
    \textbf{(2)} Rotation standardization.
    \textbf{(3)} Cropping.
    \textbf{(4)} Scale standardization.
    \textbf{(5)} Target canvas.
    \textbf{(6)} Final centered output.}
    \label{fig:standardization}
    \vspace{-5mm}
\end{figure}

\textbf{Shape Standardization. } 
To ensure consistent geometric comparison, we apply a three-stage normalization process to all silhouettes (Fig.~\ref{fig:standardization}). Given an input mask $M$, we first extract its largest connected component and compute the minimum-area bounding rectangle to determine the primary orientation. The shape is then rotated such that its longer axis aligns horizontally, where $w_{\text{box}}$ and $h_{\text{box}}$ denote the rectangle dimensions. Next, the rotated shape is scaled to fit within a $256 \times 256$ canvas using a scaling factor $\alpha = 0.9$ to preserve a uniform margin, and then centered. Finally, to address the 180° ambiguity in orientation alignment, we enforce a canonical pose. We compute the centroid's $x$-coordinate $c_x$ (from image moments) and horizontally flip the shape if $c_x < W/2$, thus ensuring its center of mass always lies in the right half of the canvas.

\textbf{Shape Matching. } Our shape matching process begins by computing SC descriptors \cite{Belongie:2002:ShaepContext} for each standardized silhouette. Similarly to the work of Belongie et al. \cite{Belongie:2002:ShaepContext}, we uniformly sample 100 contour points, a choice that balances representational detail with computational efficiency. For each point $p_i$, its descriptor $h_i$ is defined as a log-polar histogram capturing the spatial distribution of all other contour points, using the standard configuration of 12 angular and 5 logarithmic radial bins.

To compare a query shape $Q$ with a candidate $C$, we establish an optimal one-to-one correspondence between their contour point sets, $p_i^Q$ and $p_j^C$. The cost of matching two points is measured by the chi-squared distance between their descriptors, $d_{\chi^2}(h_i^Q, h_j^C)$. The overall shape similarity is then defined as the minimum assignment cost: $D(Q, C) = \min_{\pi \in \Pi} \sum_{i=1}^{N} d_{\chi^2}\left(h_i^Q, h_{\pi(i)}^C\right),$ where $\pi$ denotes a permutation in the assignment space $\Pi$ and $N$ is the number of sampled contour points. This optimization is solved efficiently using the Hungarian algorithm, ensuring globally optimal point correspondences.

\textbf{Geometric Re-ranking. } We employ a geometric verification stage for further re-ranking the top-10 candidates. This step eliminates matches whose point correspondences are geometrically inconsistent. For each candidate, a RANSAC-based estimator is used to compute the optimal affine transformation that aligns the matched points. The final selection is determined by geometric stability, defined as the lowest mean re-projection error across the inlier correspondences. This ensures that retrieved references are not only visually similar but also structurally coherent, providing reliable guidance for subsequent generation.

\textbf{Inverse Transformation for Alignment. } The final step aligns the high-resolution retrieved RGB image $R$ and its mask $M$ to the coordinate space of the original input query $Q$. Let $A_1$ denote the affine transformation that maps the retrieved mask $M$ to its standardized form $M'$, and let $A_2$ denote the transformation that maps the query $Q$ to its standardized form $Q'$. From the re-ranking step, we obtain a transformation $T$ that aligns $M'$ with $Q'$. The composite transformation $T_{\text{final}}$, which maps the original retrieved image space to the original query image space, is defined as $T_{\text{final}} = A_2^{-1} \cdot T \cdot A_1$. Applying $T_{\text{final}}$ to the retrieved image $R$ and mask $M$ yields the aligned outputs $R''$ and $M''$, which are spatially coherent with the input query $Q$.

\begin{figure}[!t]
 \centering
 \includegraphics[width=\textwidth]{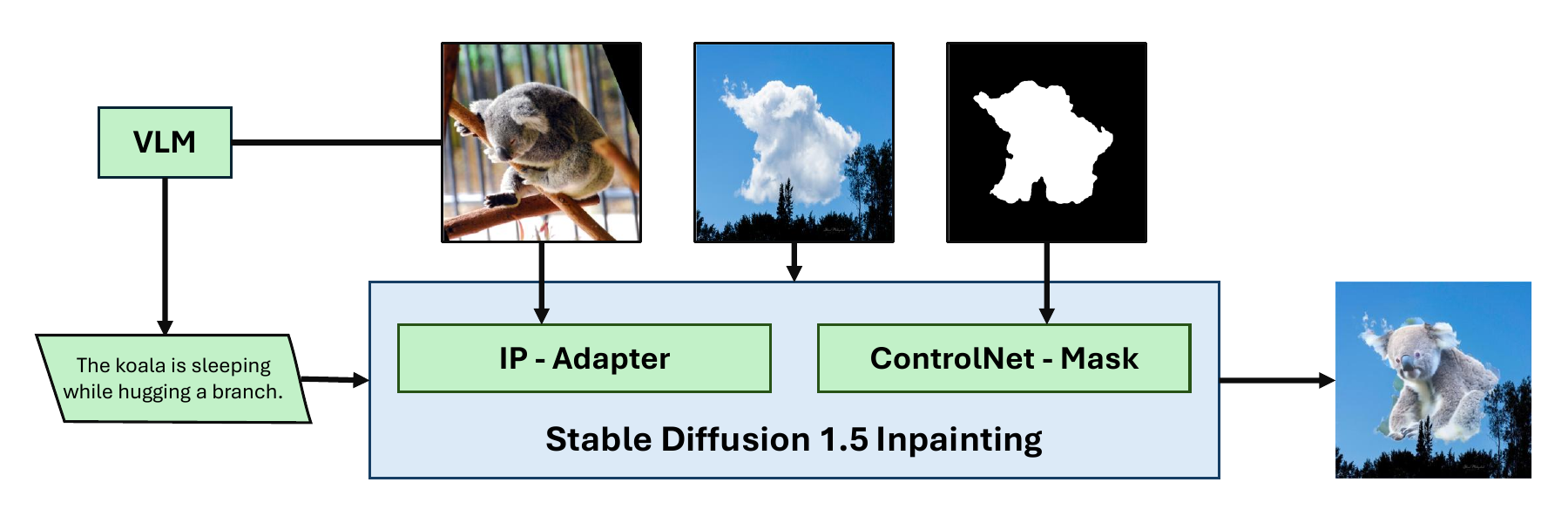}
 \vspace{-8mm}
 \caption{Reference-Guided generation pipeline.}
 \label{fig:cb-gen}
 \vspace{-5mm}
\end{figure}

\textbf{Reference-Guided Generation. } As shown in Fig.~\ref{fig:cb-gen}, the transformation process begins by retrieving a segmentation mask, its category label, and a visually similar reference image from the query results. The final image is then synthesized with Stable Diffusion 1.5 inpainting \cite{Rombach_2022_CVPR}, guided by three complementary inputs. ControlNet \cite{zhang2023addingconditionalcontroltexttoimage} uses the input mask to preserve the target silhouette, ensuring that the generated output conforms to the original shape. In parallel, IP-Adapter \cite{Ye:2023:IPAdapter} transfers visual details from the reference image, including colors, textures, and overall style. To further align the result with the intended subject, a concise textual description of the animal is automatically generated by Gemini Flash 2.5 and incorporated into the prompt.

\textbf{Image Blending. } In the final stage, the generated image $I_{\text{gen}}$ is integrated with the original input $I_{\text{orig}}$ to preserve both the object silhouette and the surrounding background. We apply soft blending within the foreground mask $M$ by interpolating $I_{\text{gen}}$ and $I_{\text{orig}}$ with a weighting factor $\alpha \in [0,1]$, while retaining the original background outside the mask. The final output $I_{\text{final}}$ is computed as:

\begin{equation}
I_{\text{final}} = \alpha \cdot (M \odot I_{\text{gen}}) + (1 - \alpha) \cdot (M \odot I_{\text{orig}}) + (1 - M) \odot I_{\text{orig}},
\end{equation}
where $\odot$ denotes element-wise multiplication. In our experiments, we set $\alpha = 0.5$ to achieve a balanced integration of generated content and original detail.

\vspace{-5mm}

\section{Animal Silhouette Corpus}

\vspace{-2mm}

Our methods rely on a large, high-quality corpus of animal shapes. To meet this need, we constructed the \textbf{Animal Silhouette Corpus}, curated from the Open Images V7 dataset \cite{OpenImages}. Its construction followed a multi-stage pipeline to ensure each silhouette is both clean and representative of an animal’s true form.

\begin{figure}[!t]
    \centering
    \begin{subfigure}[b]{0.255\textwidth}
        \includegraphics[width=\linewidth]{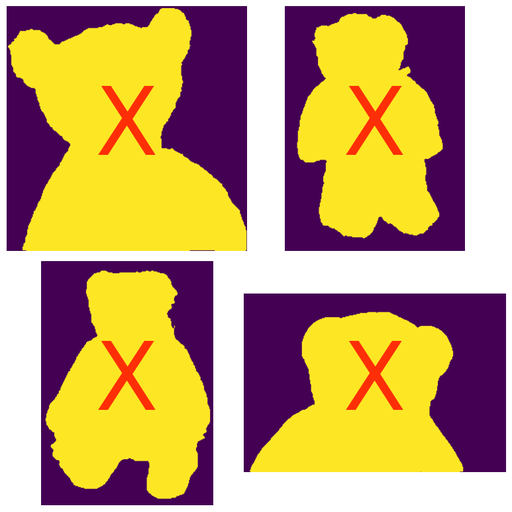}
        \caption{Semantic Filtering}
        \label{fig:filter_semantic}
    \end{subfigure}
    \hfill
    \begin{subfigure}[b]{0.25\textwidth}
        \includegraphics[width=\linewidth]{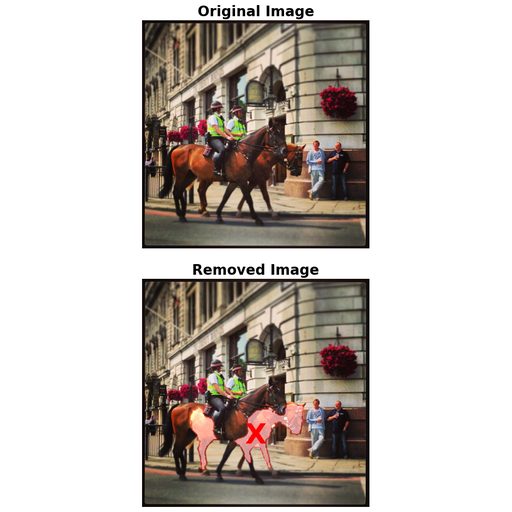}
        \caption{Contour Integrity}
        \label{fig:filter_integrity}
    \end{subfigure}
    \hfill
    \begin{subfigure}[b]{0.22\textwidth}
        \includegraphics[width=\linewidth]{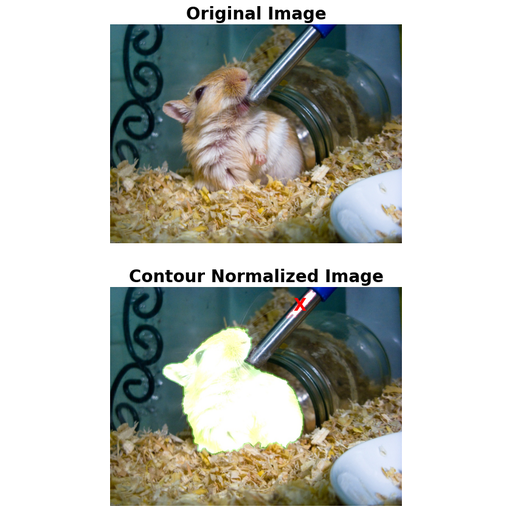}
        \caption{Normalization}
        \label{fig:filter_normalization}
    \end{subfigure}
    \hfill
    \begin{subfigure}[b]{0.25\textwidth}
        \includegraphics[width=\linewidth]{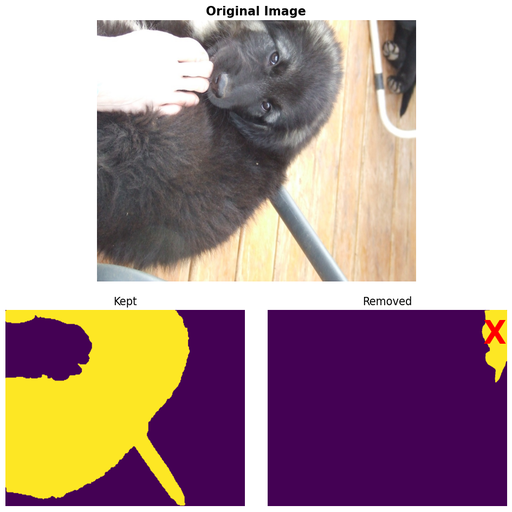}
        \caption{Mask Disparity}
        \label{fig:filter_disparity}
    \end{subfigure}
    \caption{Key filtering steps in our curation pipeline: semantic validation, contour integrity, normalization, and outlier removal.}
    \label{fig:filtering_pipeline_overview}
    \vspace{-5mm}
\end{figure}

\textbf{Source and Semantic Filtering. }
We began with 102,062 instance segmentation masks from the \texttt{Animal} superclass in Open Images V7~\cite{OpenImagesSegmentation}. For semantic integrity, we removed \texttt{Teddy bear} annotations (Fig.~\ref{fig:filter_semantic}), leaving 101,014 masks.

\textbf{Contour Integrity and Normalization. }
Each mask was required to represent a contiguous silhouette. Many were fragmented (e.g., animals behind fences) or noisy. We retained a mask $M$ only if its largest contour $C_{\max}$ covered at least 95\% of its area. Invalid masks were discarded, and valid ones normalized by keeping only $C_{\max}$ (Fig.~\ref{fig:filter_normalization}). This removed 8,118 masks, leaving 92,896 instances.


\textbf{Data-driven Thresholds. } 
To guide filtering, we analyzed Mask Area Disparity and IoU distributions. Both revealed outliers and overlaps but lacked clear cutoffs, so thresholds were set by manual inspection of 5,000 samples.

\textbf{Outlier Filtering. }
Masks smaller than one-fifth the largest in-class instance were discarded, removing 6,792 annotations (Fig.~\ref{fig:filter_disparity}), leaving 86,104 masks.

\textbf{Overlap and Duplication Removal. }
To handle overlaps, we applied IoU-based rules:  
\begin{itemize}
    \item \textit{Duplicates ($IoU > 0.9$):} retain larger mask.  
    \item \textit{Occlusions ($0.6 < IoU \leq 0.9$):} discard larger mask.  
    \item \textit{Ambiguous ($0.2 < IoU \leq 0.6$):} discard both.  
    \item \textit{Containment ($IoU \leq 0.2$):} remove $A$ if $\text{Area}(I)/\text{Area}(A) > 0.7$.  
\end{itemize}
This resolved 9,145 conflicts, yielding 76,959 validated masks.

\textbf{Class Balancing and Final Corpus. }  
Despite filtering, class imbalance remained, with counts ranging from 23 to 15,376. To normalize representation, we capped each class at 500, sampling where necessary. The final {Animal Silhouette Corpus} contains \textbf{28,586 silhouettes across 72 species}, serving as the foundation for our method.

\vspace{-3mm}

\section{Ablation Study}

\vspace{-3mm}

\subsection{Geometry-based Retrieval Evaluation}
We conducted experiments to evaluate the effectiveness of geometric components used in different stages of our retrieval process.  
\begin{table}[t!]
\centering
\caption{Comparison of initial shape retrieval methods. Metrics are averaged over all query–retrieved pairs.}
\label{tab:retrieval_comparison}
\begin{tabular}{@{}lcccc@{}}
\toprule
\textbf{Method} & \textbf{Inlier Ratio} & \textbf{Residual Error}  & \textbf{IoU} & \textbf{Avg. Time (s)} \\
\midrule
HSC~\cite{Wang:2014:HSC} & 0.3030 & 69.49 & 0.4181 & \textbf{4.9} \\
IDSC + DP~\cite{Ling:2007:IDSC} & 0.3124 & 63.63 & 0.4143 & 62.6 \\
SC + DP~\cite{Belongie:2002:ShaepContext} (Ours) & \textbf{0.3168} & \textbf{60.82} & \textbf{0.5144} & 59.3 \\
\bottomrule
\end{tabular}
\vspace{-5mm}
\end{table}

\textbf{Initial Retrieval}  
We compared our chosen initial retrieval method (SC + DP~\cite{Belongie:2002:ShaepContext}) against two alternatives: IDSC + DP~\cite{Ling:2007:IDSC} and the speed-oriented HSC~\cite{Wang:2014:HSC}, using the full corpus. As shown in Table~\ref{tab:retrieval_comparison}, HSC achieved the fastest runtime but performed worst in terms of geometric fidelity, making it unsuitable for providing high-quality visual references. The choice between SC and IDSC was more subtle. Our animal corpus spans a broad range of shapes, from highly articulated to simple forms, while the query silhouettes are almost exclusively non-articulated. In this cross-domain setting, SC’s general-purpose descriptors proved more effective, providing consistent similarity measures across the diverse corpus. The results confirm this, with SC achieving the best performance across all metrics, making it the clear choice for our framework.

\textbf{Re-ranking}  
In the re-ranking stage, we evaluated our RANSAC-based geometric verification against a learning-based point cloud registration method (i.e., LP~\cite{Bai:2010:LP}). Results in Table~\ref{tab:reranking_comparison} show that while LP achieved a higher inlier ratio, our RANSAC approach produced a substantially lower residual error and much higher post-alignment IoU. These results demonstrate that RANSAC provides more precise geometric alignment, which is essential for the Visual-RAG pipeline.  

\begin{table}[t!]
\centering
\caption{Performance comparison of re-ranking methods.}
\label{tab:reranking_comparison}
\begin{tabular}{@{}lcccc@{}}
\toprule
\textbf{Method} & \textbf{Inlier Ratio} & \textbf{Residual Error}  & \textbf{IoU} & \textbf{Avg. Time (s)} \\
\midrule
SC + DP + LP & \textbf{0.5374} & 23.0210 & 0.1241 & 2.394 \\
SC + DP + RANSAC (Ours) & 0.4527 & \textbf{0.4307} & \textbf{0.2761} & \textbf{2.262} \\
\bottomrule
\end{tabular}
\vspace{-5mm}
\end{table}

\textbf{Comparison with O2O Framework}  
We also compared our complete pipeline against the state-of-the-art O2O framework~\cite{Zheng2019:O2O}. This experiment was conducted on a representative subset of 2,153 images, created by sampling at most 30 images per class, since building the O2O offline index for the full corpus of 28,586 images was computationally prohibitive. As shown in Table~\ref{tab:o2o_comparison}, O2O achieved substantially faster retrieval, with an average time of 0.224 seconds. However, our method outperformed O2O across all geometric fidelity metrics, achieving a much higher post-alignment IoU (54.44\% vs.\ 31.87\%) and lower residual error. For our approach, where the quality of a single retrieved instance is critical for guiding creative generation, this accuracy gain justifies the additional computational cost.  

\begin{table}[t!]
\centering
\caption{Comparison of our full retrieval pipeline with the O2O framework on a subset of 2,153 images.}
\label{tab:o2o_comparison}
\begin{tabular}{@{}lcccc@{}}
\toprule
\textbf{Method} & \textbf{Inlier Ratio} & \textbf{Residual Error} & \textbf{IoU} & \textbf{Avg. Time (s)} \\
\midrule
O2O-r + SC + HSC + LP~\cite{Zheng2019:O2O} & 30.62 & 77.29 & 31.87 & \textbf{0.224} \\
SC + DP + RANSAC (Ours) & \textbf{34.68} & \textbf{56.82} & \textbf{54.44} &  4.43 \\
\bottomrule
\end{tabular}
\end{table}

\begin{figure}[!t]
    \centering
    \begin{subfigure}[b]{0.48\textwidth}
        \includegraphics[width=\linewidth]{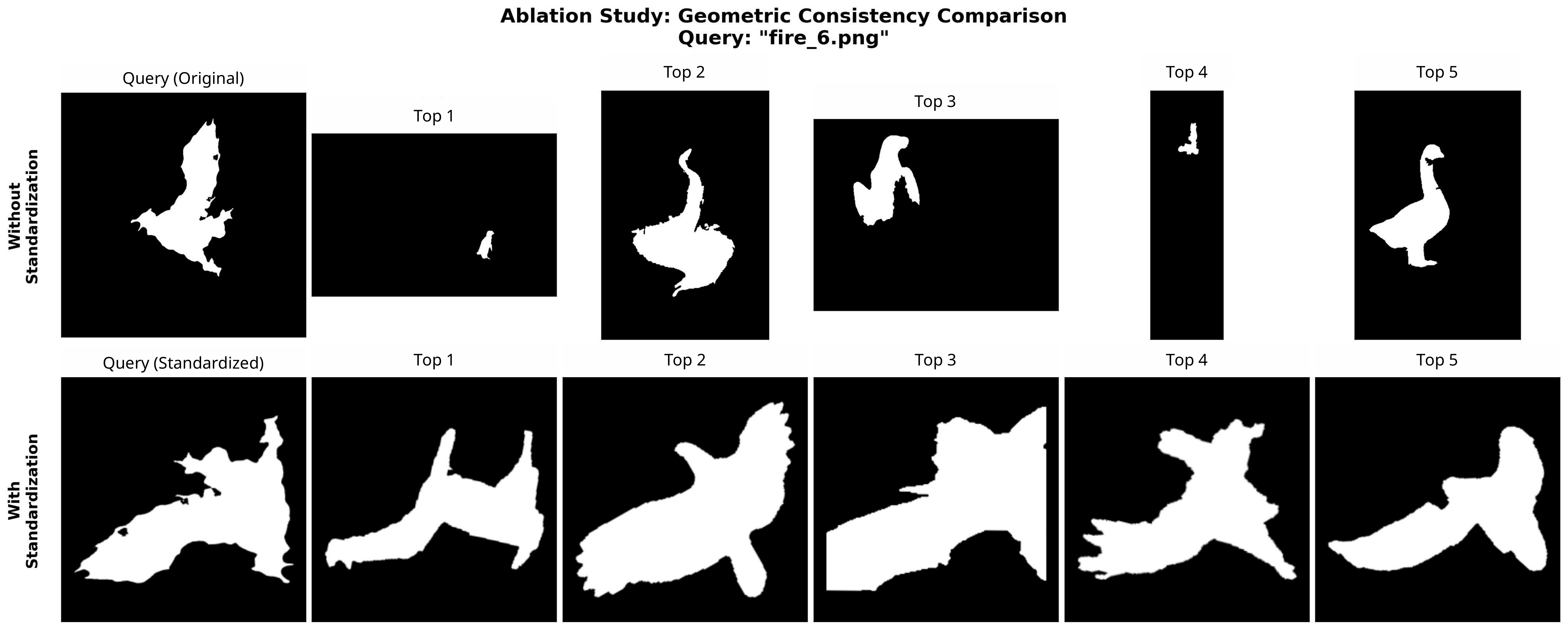}
        \label{fig:comparison_standardization_fire_6}
    \end{subfigure}
    \hfill
    \begin{subfigure}[b]{0.48\textwidth}
        \includegraphics[width=\linewidth]{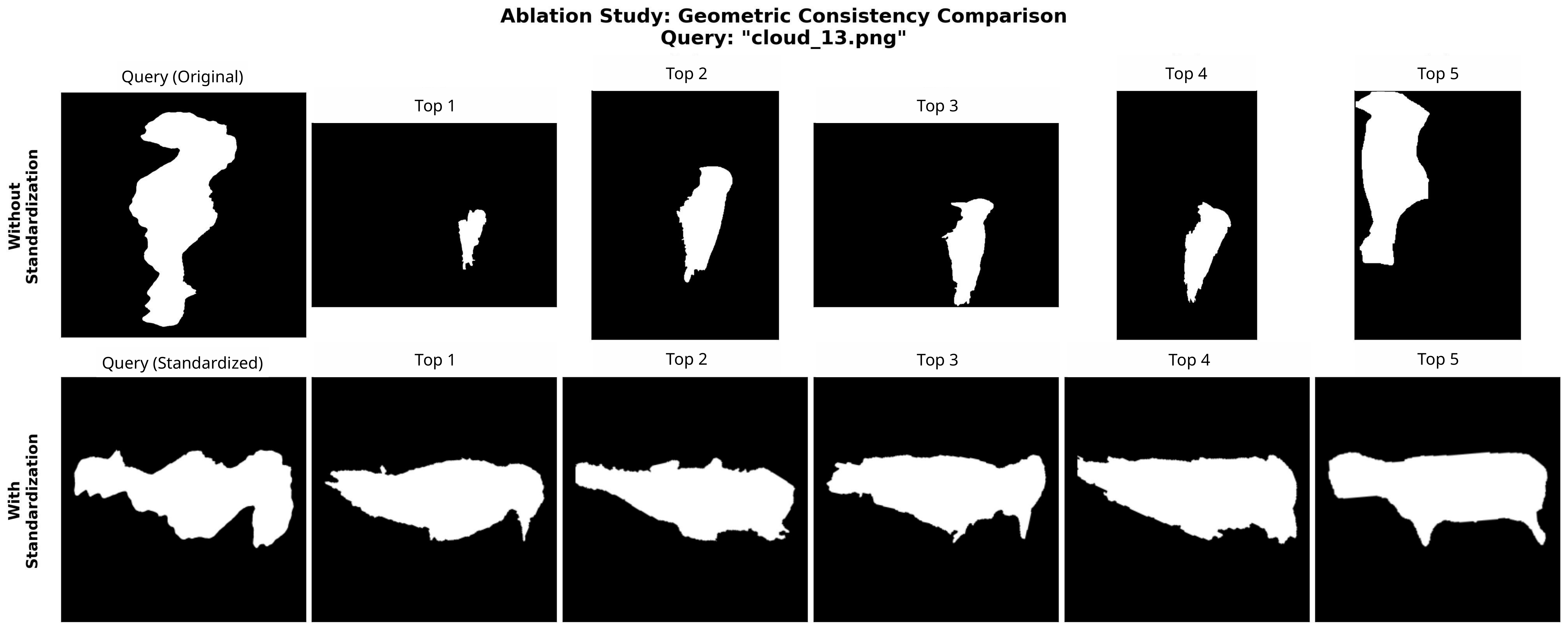}
        \label{fig:comparison_standardization_cloud_13}
    \end{subfigure}
    \hfill
    \vspace{-5mm}
    \caption{Visualize top 5 results from retrieval by SC + DP + RANSAC, with and without shape standardization.}
    \label{fig:comparison_standardization}
    \vspace{-5mm}
\end{figure}

\vspace{-5mm}

\subsection{Effectiveness of Shape Standardization}

\vspace{-2mm}

The importance of our shape standardization module is demonstrated in Fig.~\ref{fig:comparison_standardization}. Without a canonical representation for scale, rotation, and translation, the retrieval process produces geometrically inconsistent matches, especially for complex or articulated silhouettes. In such cases, the retrieved shapes fail to preserve the structural essence of the query, as reflected in the qualitative examples. Quantitatively, the average inlier ratio without standardization drops to just {13.4\%}, underscoring that this step is indispensable for achieving high-quality retrieval and reliable downstream geometric alignment.

\vspace{-5mm}

\subsection{Comparison with Naive Diffusion Baseline}

\vspace{-2mm}



\begin{figure}[!t]
    \centering
    \includegraphics[width=0.7\textwidth]{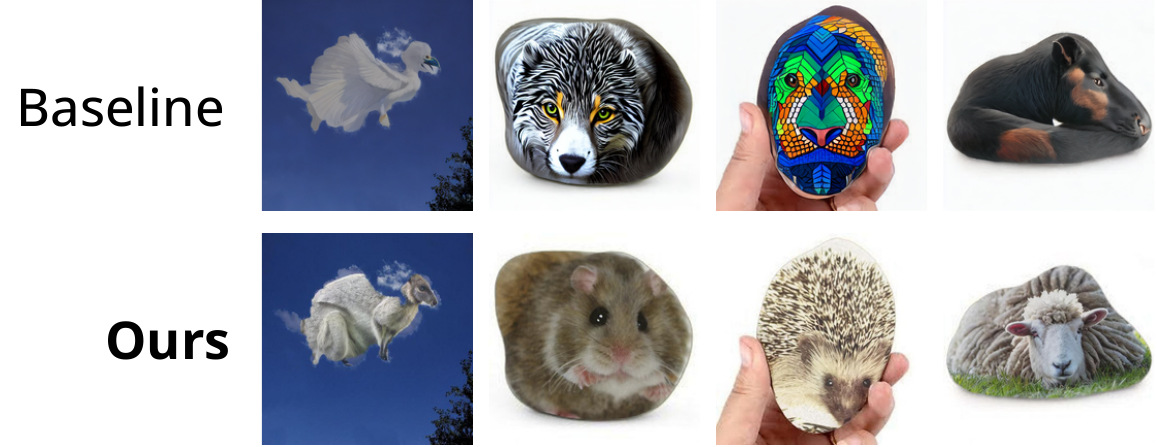}
    \caption{Comparison with the Naive ControlNet baseline (using a generic "animal" prompt).}
    \label{fig:baseline_comparison}
\end{figure}

To validate the necessity of our structured Visual-RAG framework, we compared it with a naive diffusion baseline that combines a ControlNet-Canny model \cite{zhang2023addingconditionalcontroltexttoimage} and a generic text prompt (e.g., “an artistic painting of an animal”). Using the same Stable Diffusion 1.5 inpainting setup, this baseline adheres to the silhouette but merely hallucinates random animals that fit the shape. As shown in Figure~\ref{fig:baseline_comparison}, its outputs are inconsistent, lack creative control, and often fail to produce coherent results.

\vspace{-3mm}

\section{User Study}\label{sec:experiments}

\vspace{-3mm}

\subsection{Experimental Design}

\begin{figure}[!t]
  \centering
  \includegraphics[width=\textwidth, height=3.5cm]{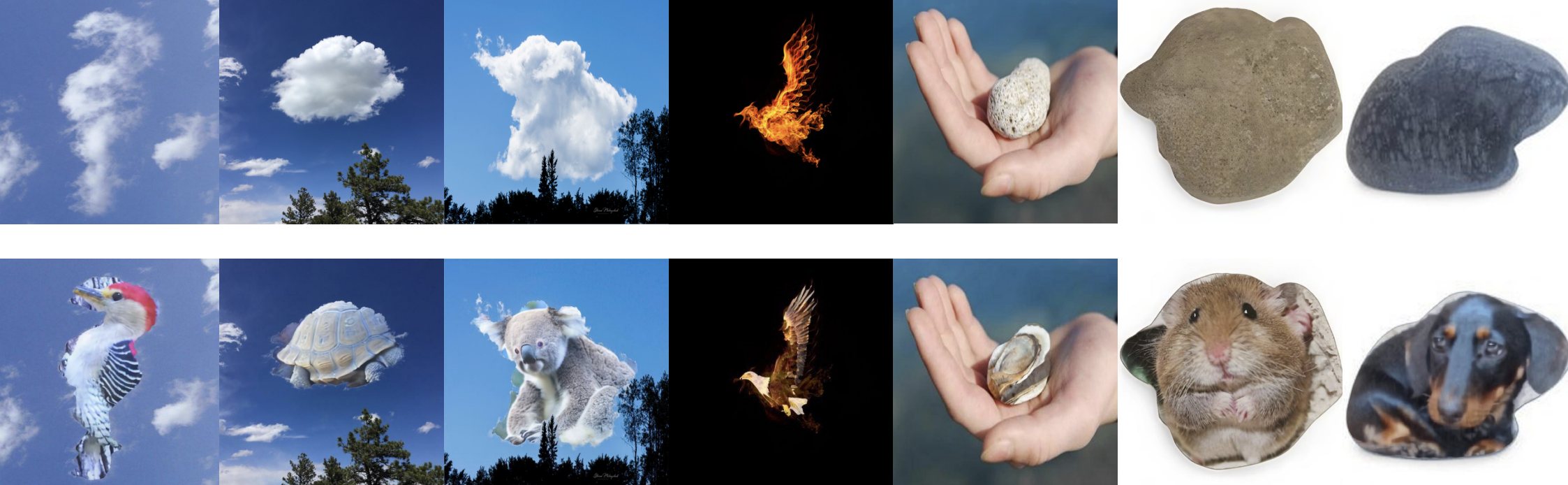}
  \vspace{-5mm}
  \caption{Sample results from the user study. Each input image (top row) is paired with the corresponding outputs generated by Visual-RAG (bottom row).}
  \label{fig:visualize}
  \vspace{-5mm}
\end{figure}

To evaluate Visual-RAG, we selected 40 ambiguous natural images, including stones, clouds, fire, and leaves. Each input was processed to generate one output image, resulting in a test set of 40 generated samples (Fig. \ref{fig:visualize}). Twelve participants, drawn from both technical and non-technical backgrounds, rated each output on a 5-point Likert scale across four dimensions: Aesthetics (visual appeal), Shape Fitness (silhouette adherence), Impression (creative impact), and Overall Quality. To mitigate bias, images were shown in randomized order, and participants were not informed about the underlying method. 

\vspace{-3mm}

\subsection{Results}

\begin{table}[t!]
\begin{center}
\caption{Mean scores and standard deviations from the user study. The highest score for each metric is highlighted in bold.}
\label{tab:user_study_results}
\begin{tabular}{l>{\centering\arraybackslash}p{2cm}>{\centering\arraybackslash}p{2.5cm}>{\centering\arraybackslash}p{2cm}>{\centering\arraybackslash}p{2cm}}
\toprule
\textbf{Method} & \textbf{Aesthetics} & \textbf{Shape Fitness} & \textbf{Impression} & \textbf{Overall} \\
\midrule
Visual-RAG & 2.46 $\pm$ 1.23 & 2.62 $\pm$ 1.29 & 2.46 $\pm$ 1.29 & 2.51 $\pm$ 1.27 \\
\bottomrule
\end{tabular}
\end{center}
\vspace{-8mm}
\end{table}

As summarized in Table \ref{tab:user_study_results}, Visual-RAG’s performance consistently fell below the neutral midpoint of 3.0 across all metrics, with mean ratings of 2.46 for aesthetics, 2.62 for shape fitness, and 2.46 for impression. These results suggest that while the framework can produce coherent outputs, its creative impact remains limited. In particular, participants noted frequent failures where retrieved exemplars were incomplete or ambiguous, leading to outputs that lacked visual coherence. Our study revealed a strong positive correlation $(r \approx 0.75)$ between shape fitness and aesthetics, underscoring that strict adherence to the source silhouette is a key determinant of perceived quality. This finding suggests that the psychological illusion of pareidolia depends critically on structural fidelity:
when the generated image conforms closely to the input shape, viewers experience the creative discovery of “seeing” an animal within it; when fidelity breaks down, the illusion collapses, and the result is perceived as an unrelated object superimposed on the silhouette. Improving structural alignment in future iterations of Visual-RAG is therefore central to enhancing both creative impact and perceptual quality.

\vspace{-3mm}

\section{Conclusion}\label{sec:conclusion}

\vspace{-3mm}

In this paper, we proposed a framework for generating animal art from abstract silhouettes, directly tackling the challenge of computational pareidolia. Our experiments demonstrate that retrieval-based conditioning can produce visually coherent and shape-faithful generations, while a user study highlights the importance of silhouette adherence for aesthetic appeal. Future work should move beyond single exemplars by developing adaptive strategies for incomplete or ambiguous references, expanding the 72-class dictionary to foster creativity, and advancing shape decomposition and retrieval methods to improve stability and imagination.


\begin{credits}
\subsubsection{\ackname}
This research is funded by Vietnam National Foundation for Science and Technology Development (NAFOSTED) under Grant Number 102.05-2023.31. This research used the GPUs provided by the Intelligent Systems Lab at the Faculty of Information Technology, University of Science, VNU-HCM.
\end{credits}

\bibliographystyle{splncs04}
\bibliography{camera_ready_short}

\end{document}